\documentclass{article}

\usepackage{PRIMEarxiv}

\usepackage[utf8]{inputenc} 
\usepackage[T1]{fontenc}    
\usepackage{hyperref}       
\usepackage{url}            
\usepackage{booktabs}       
\usepackage{amsfonts}       
\usepackage{nicefrac}       
\usepackage{microtype}      
\usepackage{lipsum}
\usepackage{fancyhdr}       
\usepackage{graphicx}       
\usepackage{multirow} 
\usepackage{amssymb}
\usepackage{graphicx}
\usepackage{color}
\usepackage{soul}
\usepackage{newfloat}
\usepackage{listings}
\usepackage{bm}
\usepackage{bibentry}
\usepackage{microtype}
\usepackage{tabularx}
\graphicspath{{media/}}     

\title{UIFormer: A Unified Transformer-based Framework for Incremental Few-Shot Object Detection and Instance Segmentation}

\author{
  Chengyuan Zhang  \\
  Hunan University \\
  \texttt{cyzhangcse@hnu.edu.cn} \\
  \And
  Yilin Zhang  \\
  Hunan University \\
  \texttt{zhangyilin@hnu.edu.cn} \\
   \And
  Lei Zhu \\
  Hunan Agricultural University \\
  \texttt{leizhu@hunau.edu.cn} \\
  \And
  Deyin Liu \\
  Anhui University \\
  \texttt{Iedyzzu@outlook.com} \\
  \And
  Lin Wu \\
  Swansea University \\
  \texttt{l.y.wu@swansea.ac.uk} \\
  \And
  Bo Li \\
  Northwestern Polytechnical University \\
  \texttt{libo@nwpu.edu.cn} \\
  \And
  Shichao Zhang \\
  Guangxi Normal University\\
  \texttt{zhangsc@mailbox.gxnu.edu.cn}\\
  \And
  Mohammed Bennamoun\\
  University of Western Australia \\
  \texttt{mohammed.bennamoun@uwa.edu.au}
  \And
  Farid Boussaid \\
  University of Western Australia \\
  \texttt{farid.boussaid@uwa.edu.au}
}

\begin{document}
\maketitle

\begin{abstract}
This paper introduces a novel framework for unified incremental few-shot object detection (iFSOD) and instance segmentation (iFSIS) using the Transformer architecture. Our goal is to create an optimal solution for situations where only a few examples of novel object classes are available, with no access to training data for base or old classes, while maintaining high performance across both base and novel classes. To achieve this, We extend Mask-DINO into a two-stage incremental learning framework. Stage 1 focuses on optimizing the model using the base dataset, while Stage 2 involves fine-tuning the model on novel classes. Besides, we incorporate a classifier selection strategy that assigns appropriate classifiers to the encoder and decoder according to their distinct functions. Empirical evidence indicates that this approach effectively mitigates the over-fitting on novel classes learning. Furthermore, we implement knowledge distillation to prevent catastrophic forgetting of base classes. Comprehensive evaluations on the COCO and LVIS datasets for both iFSIS and iFSOD tasks demonstrate that our method significantly outperforms state-of-the-art approaches.
\end{abstract}

\section{Introduction}

In the field of computer vision, incremental few-shot object detection (iFSOD) and instance segmentation (iFSIS) stand as critical tasks. iFSOD focuses on detecting novel object classes incrementally with only a few examples, while simultaneously preserving knowledge of previously learned classes without needing to re-access data from these base classes. iFSIS extends this challenge to a more granular level by assigning categorical labels at the pixel level, enhancing the model's capability for fine-grained segmentation. 

So far, significant advancements have been made in both iFSOD and iFSIS. The ONCE framework \cite{perez2020incremental} was the first to address the iFSOD problem, but its basic learning strategy struggles to retain knowledge of base classes when novel classes are introduced. To combat catastrophic forgetting, methods such as LEAST \cite{li2021class} and Incremental-DETR \cite{dong2023incremental} have employed transfer learning and knowledge distillation, respectively, to safeguard the knowledge of base classes. Sylph \cite{yin2022sylph} further mitigates knowledge forgetting by using a base detector with class-agnostic localization. For the more challenging iFSIS tasks, iMTFA \cite{ganea2021incremental} calculates the class embedding averages using cosine similarity, allowing the model to extend to novel classes without additional training.

Recently, a unified framework, iFS-RCNN \cite{nguyen2022ifs}, has been developed to bridge these tasks by adapting the general Mask R-CNN model. This model introduces a probit-based object classifier and an uncertainty-guided bounding box predictor, delivering robust performance in both iFSOD and iFSIS. Despite the clear distinctions between region-level and pixel-level tasks, the inherent adaptability of Mask R-CNN supports seamless task integration. However, CNN-based models generally lag behind transformer-based models in handling complex semantics, particularly in large-scale datasets. This gap underscores the pressing need for transformer-based solutions for iFSOD and iFSIS in data-intensive scenarios.

\subsubsection{Motivation} 

Currently, transformer-based models for incremental object detection and instance segmentation remain disjointed. This lack of integration restricts data sharing and cooperative learning opportunities across the tasks. Mask-DINO \cite{li2023mask} , a transformer-based extension of DINO \cite{zhang2022dino}, illustrates the potential for unifying region-level and pixel-level feature representations, suggesting that detection and segmentation can mutually enhance each other within a single transformer-based architecture. Nonetheless, a significant limitation observed in Mask-DINO is the uniform classifier across its encoder and decoder components, which our empirical studies suggest degrades performance in both iFSOD and iFSIS. This observation leads us to pose two pivotal questions: 

\begin{itemize}
    \item \textit{Is it feasible to unify detection and instance segmentation using a Transformer architecture in an incremental setting?}
    \item \textit{How can we effectively assign appropriate classifiers to the encoder and decoder to reflect their distinct functionalities?} 
\end{itemize}

\subsubsection{Our Method} 

To address these inquiries, we introduce \textbf{UIFormer}, a novel unified transformer-based framework designed to support both iFSOD and iFSIS. Our approach employs a dual-phase learning strategy, initially dividing the learning of base classes into pre-training and fine-tuning phases to solidify foundational knowledge. During fine-tuning, we incorporate an attention-driven pseudo ground-truth search technique to detect and label previously unidentified objects, effectively expanding the dataset's class diversity. Additionally, to combat the loss of base class knowledge, we integrate knowledge distillation into the fine-tuning process for novel classes. Simultaneously, we propose a new class-agnostic foreground predictor for the encoder, which focuses solely on distinguishing between foreground and background, enhancing token selection for subsequent processing by the decoder, which employs a cosine-similarity classifier for precise semantic discrimination.

\subsubsection{Contributions} To sum up, our contributions are four-folds as follows:
\begin{itemize}
 \item We are the first to explore the extension of transformer-based models for learning iFSOD and iFSIS tasks in a unified manner. This goal is achieved through the introduction of UIFormer, a transformer-based framework that unifies object detection and instance segmentation.
\item We propose an classifier selection strategy to assign different classifiers to encoder and decoder based on their distinct functions. In addition, a knowledge distillation method is used in novel fine-tuning stage. Empirical evidence demonstrates that our approach effectively mitigates over-fitting and catastrophic forgetting.
 \item We conduct extensive experiments on the commonly used COCO and LIVS datasets. The results demonstrate that the proposed method outperforms the state-of-the-arts approaches by a significant margin on both iFSOD and iFSIS task.
\end{itemize}

\section{Related Work}

\subsection{FSOD Methods}
Existing FSOD methods can be classified into two categories: meta-learning-based methods and fine-tuning based methods. The meta-learning based methods\cite{Fan_Zhuo_Tang_Tai_2020, Kang_Liu_Wang_Yu_Feng_Darrell_2019, Yan_Chen_Xu_Wang_Liang_Lin_2019, zhang2023meta_detr, zichen2024fine, guang2022fully} employ episodic methodology\cite{Vinyals2016matching} to train the model, so that the model can quickly adapt to novel classes with annotated samples. The fine-tuning-based methods \cite{Chen_Wang_Wang_Qiao_2022, xin2020tfa, bo2021fsce, jia2020multi} are first trained on the base classes and then fine-tune some components while freezing the remaining to allow the model to learn how to detect new objects and mitigate overfitting on limited samples of novel classes.

\subsection{FSIS Methods}
Similar to FSOD, existing FSIS methods can be also categorized into two groups: meta-learning-based methods\cite{zhibo2020fully, claudio2018one, khoi2021fapis,xiaopeng2019meta} and fine-tuning-based methods\cite{minh2024maskdiff}. Those FSIS methods usually use Mask R-CNN as backbone network.

\subsection{iFSOD Methods}
\cite{perez2020incremental} defined the iFSOD problem for the first time and proposed the ONCE model to deal with it. ONCE uses CenterNet \cite{zhou2019objects} as a backbone to learn a class-agnostic feature extractor and a per class code generator network for novel classes. Due to it does not visit the base class and only access novel class once during novel class training, it is difficult for ONCE to remember the knowledge learned in previous tasks, and its transfer capability to detect new objects is low. \cite{li2021class} proposed a novel method LEAST to achieve less forgetting and stronger transfer capability. LEAST presents a new transfer strategy that decouples class-sensitive object feature extractor to obtain stronger transfer capability, and integrate the knowledge distillation technique using a less resource-consuming approach to alleviate the catastrophic forgetting. Dong et al. \cite{dong2023incremental} introduced Incremental-DETR, which takes DETR object detector \cite{carion2020end} as a backbone, into iFSOD. The entire model is divided into two stages, and still follows base classes pre-training and novel classes fine-tuning guideline. In order to avoid catastrophic forgetting of base class knowledge, this model adopts knowledge distillation to assist the network in detecting novel classes. A hypernetwork architecture, called Sylph, was introduced by Yin et al.\cite{yin2022sylph} to address iFSOD. The backbone of Sylph is similar to ONCE. Difference as ONCE, a base detector with class-agnostic localization is trained by Sylph on abundant base class data to decouple object classification from localization. They further modify the network structure and add normalization to the predicted parameters to achieve better performance.

\subsection{iFSIS Methods}

iMTFA\cite{ganea2021incremental} is the first work to target at more challenging iFSIS task. It employs Mask-R-CNN\cite{he2017mask}  as a backbone, and adopts a two-stage training and fine-tuning strategy to train the mask R-CNN network. Unlike traditional Mask R-CNN, iMTFA repurposes the fully-connected layers at the region of interest level to train feature extractors to generate discriminative embeddings for different instances. These embedding are subsequently used as the representation for each class inside a cosine similarity classifier. The current state of the art technique for iFSIS is iFS-RCNN\cite{nguyen2022ifs}, which is developed by extending general Mask-R-CNN model. iFS-RCNN purposes a new object class classifier based on the probit function via Bayesian learning to deal with a paucity of training examples of new classes. In addition, it uses a new uncertainty-guided bounding box predictor to refine the bounding-box prediction and to appropriately weight the loss of bounding-box prediction.

\section{Methodology}
\begin{figure*}
    \centering
    \includegraphics[height=5cm, width=\textwidth]{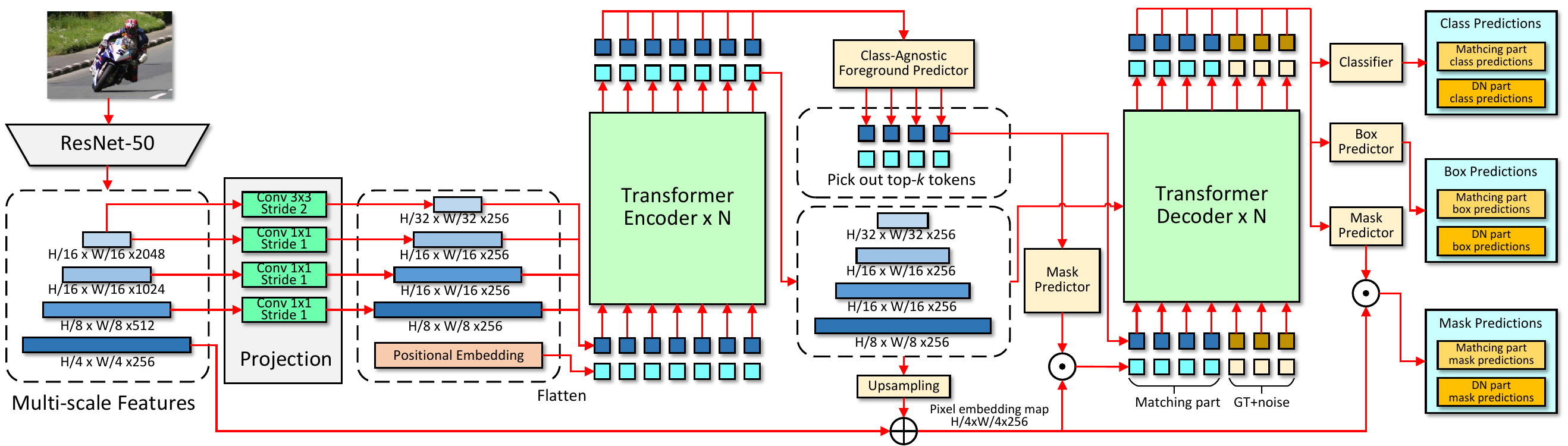}
    \caption{The backbone of UIFormer. It is a unified transformer based model to effectively align region-level task and pixel-level task. Three different prediction heads, i.e., a class-agnostic foreground predictor, a mask predictor and a box predictor are plugged on the top of the encoder to guide the model learning under foreground prediction, mask prediction, bounding box prediction.
    }
    \label{fig:example}
\end{figure*}
\subsection{Problem Definition}
Let $\mathcal{D}=\{(I_i,Y^c_i,Y^b_i,Y^m_i)\}$ be an image dataset, in which each sample $I_i \in \mathbb{R}^{H\times W \times 3}$ has the objects' classes $Y^c_i$, ground-truth bounding boxes $Y^b_i=[x,y,w,h]$, and the binary masks $Y^m_i$. Suppose we have a base class dataset $\mathcal{D}_{base}=\{(I_i,Y^c_i,Y^b_i,Y^m_i)\}^{D}_{i=1}$ with $B$ classes $\{C_k\}^B_{k=1}$ and a novel class dataset $\mathcal{D}_{novel}=\{(I_i,Y^c_i,Y^b_i,Y^m_i)\}^{K}_{i=1}$ with $N$ classes $\{C_k\}^{B+N}_{k=B+1}$ ($N$-way $K$-shot). Under iFS learning setting, $\{C_k\}^B_{k=1} \bigcap \{C_k\}^{B+N}_{k=B+1} = \varnothing$ and the base classes are no longer accessible when novel classes are coming. Given a query image $I_q$, this work aims to detect and segment all object instances belong to all classes $B+N$ in a unified incremental learning framework. To this end, our goal is to learn $N$ novel classes from $K$ new-coming samples without forgetting the semantics from base classes.    

\subsection{Architecture of UIFormer}

\subsubsection{Overview} UIFormer is a two-stage incremental few-shot learning framework that integrates object detection and instance segmentation. In Stage I, the framework trains a base model to capture rich semantics from the base datasets while incrementally adapting to a few novel classes through fine-tuning. To address the challenge of harmonizing object detection and instance segmentation within a unified framework, we employ Mask DINO as the backbone, which effectively aligns region-level and pixel-level tasks. As shown in Fig.\ref{fig:example}, the multi-scale features of the inputs are initially transformed using a $1 \times 1$ convolutional projection layer before being fed into the encoder, which incorporates positional embeddings. The encoder is topped with three distinct prediction heads: a class-agnostic foreground predictor, a mask predictor, and a box predictor, each designed to guide the model in foreground detection, mask prediction, and bounding box estimation, respectively. Additionally, to enhance iFSOD performance with the support of segmentation tasks, we adopt a mask-enhanced anchor box initialization strategy. This approach allows bounding box predictions to be derived from high-resolution predicted masks, thereby improving detection accuracy. The decoder architecture of UIFormer mirrors that of Mask DINO in the base pre-training stage, featuring a hybrid matching module to facilitate interactions between box and mask predictions, and a unified DN module to expedite convergence.

\subsubsection{Classifier Selection} In Mask DINO, the encoder and decoder share the same classifier to enhance semantics learning of each object. However, this approach is not well-suited for incremental learning task. The underlying issue is that the use of a semantic classifier during the base training phase can lead to over-fitting of the encoder, thereby degrading performance on novel datasets. To address this limitation, we select appropriate classifiers for the encoder and decoder. In particular, we introduce a class-agnostic foreground predictor that functions as a binary classifier, discriminating between foreground and background of each image. By not considering the specific category of each object, this method effectively improves foreground discrimination across both base and novel data. As a result, the original unified query selection in Mask DINO is replaced with a strategy that selects foregrounds based on top-ranked recognition confidence. This modification has been empirically validated as a superior query selection strategy for both iFSOD and iFSIS.

\begin{figure}[tb]
    \centering
    \includegraphics[height=7.2cm]{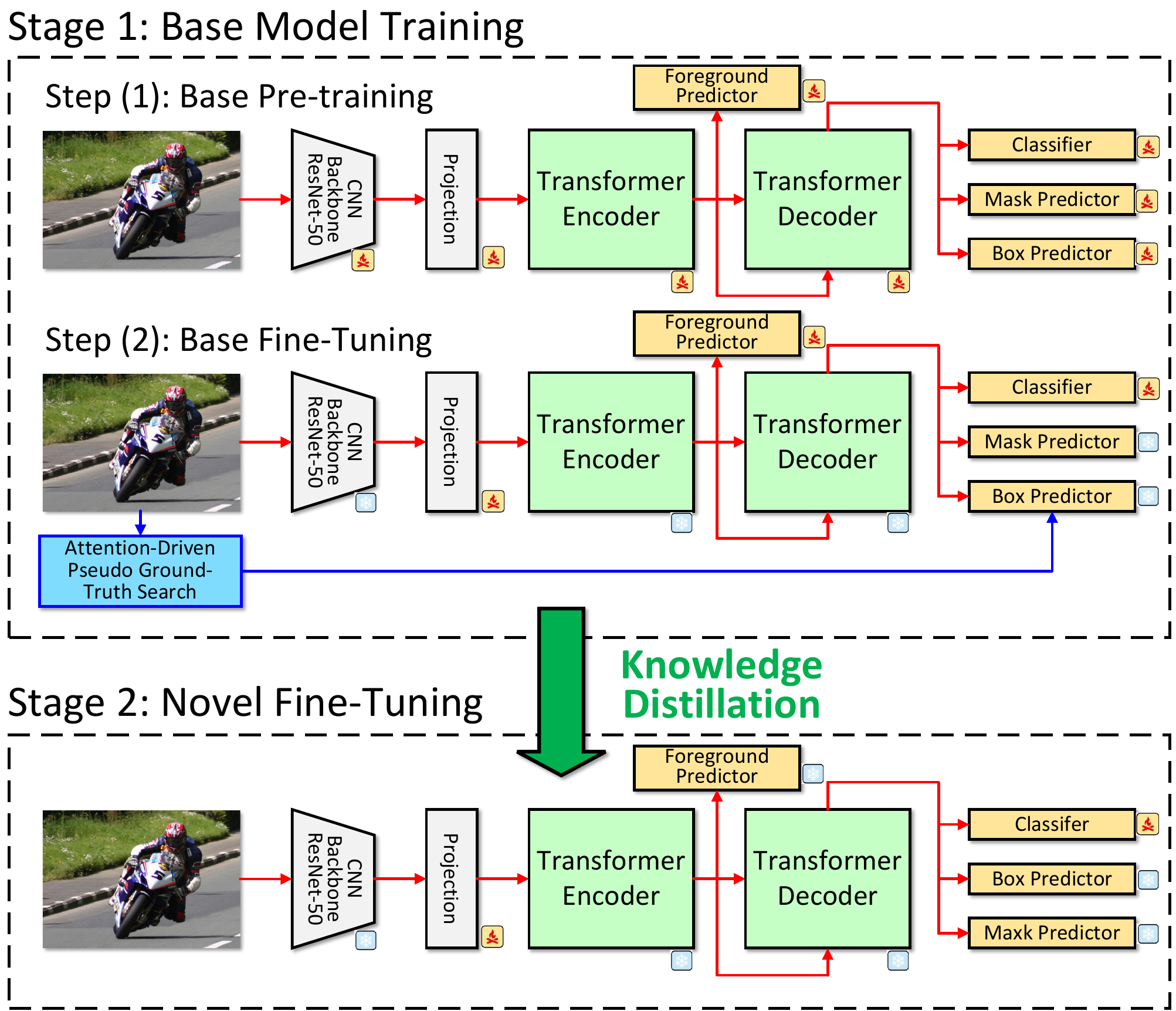}
    \caption{A two-stage training strategy. The stage 1 is base model training, which includes two step: Step (1) is to training the full model on both object detection and instance segmentation tasks; Step (2) is a fine-tuning on the base by an attention-driven pseudo-labels based self-supervised task. The stage 2 is novel fine-tuning, which further optimize the projection layer and classifier to acquire novel semantic knowledge. Knowledge distillation method is utilized to reduce the discrepancy between the outputs of the projection layer in the base model and those in the novel mode.
    }
    \label{fig:two-stage}
\end{figure}

\subsection{Stage 1: Base Model Training}
To further enhance semantic learning from base classes and alleviate over-fitting, we propose a two-step base model training strategy, as shown in fig.~\ref{fig:two-stage}. The step (1), termed base model pre-training, involves training the complete model on both object detection and instance segmentation tasks. The step (2) entails fine-tuning the base model through an attention-driven, pseudo-label-based self-supervised task.

\subsubsection{Step (1): Base Pre-Training} In this step, we pre-train the model on the base class dataset $\{C_k\}^B_{k=1}$ using both object detection and instance segmentation tasks. Along with the standard loss functions from DETR \cite{carion2020end}, we introduce three novel loss functions: $\mathcal{L}_{mask}$, $\mathcal{L}_{mask_interm}$, and $\mathcal{L}_{mask_dn}$ to enhance instance segmentation learning. Specifically, $\mathcal{L}_{mask} = \lambda_{ce}\mathcal{L}_{ce} + \lambda_{dice}\mathcal{L}_{dice}$ is the mask loss~\cite{cheng2022masked}, where $\lambda_{ce}$ and $\lambda_{dice}$ are the loss weight parameters. The total loss function of pre-training is:
\begin{equation}
    \mathcal{L}_{bpt} = \mathcal{L}_{cls} + \mathcal{L}_{mask} + \mathcal{L}_{box} + \mathcal{L}_{interm} + \mathcal{L}_{dn} + \mathcal{L}_{aux},
\end{equation}
where $\mathcal{L}_{cls}$ denotes decoder classification loss that is realized by sigmoid focal loss. $\mathcal{L}_{box} = \lambda_1\mathcal{L}_1 + \lambda_{giou}\mathcal{L}_{giou}$ is the box regression loss. $\mathcal{L}_{interm} = \mathcal{L}_{cls\_interm\_binary} + \mathcal{L}_{mask\_interm} + \mathcal{L}_{box\_interm}$ is the prediction loss of the top-$k$ output tokens of encoder. $\mathcal{L}_{dn} = \mathcal{L}_{cls\_dn} + \mathcal{L}_{mask\_dn} + \mathcal{L}_{box\_dn}$ is the denoising loss to promote accelerated convergence of the model. $\mathcal{L}_{aux} = \sum^{L}_{i=1}(\mathcal{L}^{(i)}_{cls}+\mathcal{L}^{(i)}_{box}+\mathcal{L}^{(i)}_{mask})$ is an auxiliary decoding loss that aims to train each layer of the decoder for class prediction, bounding box prediction and mask prediction, where $L$ denotes the number of layer.

\subsubsection{Step (2): Base Fine-Tuning} To further explore the semantic knowledge in base dataset, we incorporate an effective class space expansion-based self-supervised learning strategy. This approach aims to actively capture unlabeled objects and fine-tune the class-sensitive components of the backbone using pseudo ground-truth. Inspired by OW-DETR \cite{gupta2022ow}, we adopt an attention-driven pseudo ground-truth search method to identify unlabeled objects and generate their pseudo ground-truth, thereby expanding the class space. Specifically, given an input image $I_i \in \mathcal{D}_b$ with $|Y^c_i|$ labeled objects, suppose the model give $P>|Y^c_i|$ predictions, each represented as $(\hat{Y}^c_i,\hat{Y}^b_i,\hat{Y}^m_i)$. After bipartite matching, $|Y^c_i|$ out of $P$ predictions will be matched with the ground-truth labels. Meanwhile, the remaining $P - |Y^c_i|$ predictions, which are matched with no-object labels, will have the top-$k$ predictions (where $k=5$ in our experiments) selected as pseudo ground-truth based on objectness scores $s$:
\begin{equation}
    s=\frac{1}{h\times w}\sum^{x+\frac{w}{2}}_{x-\frac{w}{2}}\sum^{y+\frac{h}{2}}_{y-\frac{h}{2}}\bm{A},
\end{equation}
where $\bm{A}\in \mathbb{R}^{H\times W\times 1}$ denotes an attention map, which generating by the feature map $\bm{f} \in \mathbb{R}^{\frac{16}{H} \times \frac{16}{W} \times 1024}$ from backbone after channel pooling and up-sampling. Subsequently, the parameters of the CNN feature extractor, transformer-based encoder and decoder, as well as the box predictor and mask predictor, are frozen. The projection model, foreground predictor, and classifier are then fine-tuned within the expanded class space. Notably, we identified issues caused by using pseudo boxes and pseudo masks during fine-tuning; therefore, we only utilize pseudo class labels for fine-tuning the classifier. Another significant finding is that denoising training is no longer required after pre-training, as accelerating convergence becomes unnecessary. In summary, the total loss function for base fine-tuning is as follows:
\begin{equation}
    \mathcal{L}_{bft} = \mathcal{L}_{pcls}+\mathcal{L}_{mask}+\mathcal{L}_{box}+\mathcal{L}_{interm}+\mathcal{L}_{aux},
\end{equation}
where $\mathcal{L}_{pcls}$ denotes classification loss after class space expansion.

\subsection{Stage 2: Novel Model Fine-Tuning}
To effectively capture novel semantics from a few new samples while preserving the knowledge acquired from base classes, we propose a knowledge distillation-based fine-tuning strategy. Before this stage, we initialize the novel model using parameters from the base model. Then, we fine-tune the projection layer and classifier to acquire novel semantic knowledge, while freezing the CNN feature extractor, encoder, decoder, and the three prediction heads to prevent catastrophic forgetting. Additionally, to mitigate knowledge loss during fine-tuning of the projection layer, we introduce knowledge distillation to reduce the discrepancy between the outputs of the projection layer in the base model and those in the novel model. To account for multi-scale features in the input image, the distillation loss function is defined as follows:
\begin{equation}
    \mathcal{L}^{kd}_{Proj}=\sum^{l}_{i=1}\frac{1}{2N^{novel}_{i}}\sum^{w}_{j=1}\sum^{h}_{k=1}\sum^{c}_{n=1}(1-Y^m_{ijk})||\bm{f}^{novel}_{ijkn}-\bm{f}^{base}_{ijkn}||^2,
\end{equation}
\begin{equation}
    N^{novel}_{i} = \sum^{w}_{j=1}\sum^{h}_{k=1}(1-Y^m_{ijk}),
\end{equation}
where $l$ denotes the number of multi-scale feature maps. $w$, $h$, and $c$ represent the width, height, and number of channels in a feature map, respectively. $Y^m_{ijk}$ is the binary mask corresponding to novel classes in the $i$-th scale feature map. $\bm{f}^{novel}_{ijkn}$ and $\bm{f}^{base}_{ijkn}$ refer to the outputs of the $i$-th feature map from the projection layer in the novel and base models, respectively. The overall loss function for this stage is defined as follows:
\begin{equation}
    \mathcal{L}_{nft} = \mathcal{L}_{cls} + \mathcal{L}_{mask} + \mathcal{L}_{box} + \mathcal{L}_{interm} + \mathcal{L}_{aux} + \mathcal{L}^{kd}_{proj}. 
\end{equation}
Note that, we still abandon the denoising training in this stage due to the same reason in base model fine-tuning.

\section{Experiments}

\subsection{Experimental Setup}

\begin{table}[]
  \caption{\normalsize  Comparison of iFSOD and iFSIS with $k$=10 for different classification heads combination strategies on COCO 2014. We report AP and AP50 for base and novel classes. Optimal results are in \textbf{bold}, sub-optimal best results are in \ul{underline}. }
  \centering
  \renewcommand{\arraystretch}{1.3} %
  \resizebox{0.7\columnwidth}{!}{
  \begin{tabular}{|c|cccc|cccc|}
\hline
\multirow{3}{*}{Method} & \multicolumn{4}{c|}{Object Detection}                       & \multicolumn{4}{c|}{Instance Segmentation}                 \\ \cline{2-9} 
                        & \multicolumn{2}{c|}{Base}      & \multicolumn{2}{c|}{Novel} & \multicolumn{2}{c|}{Base}      & \multicolumn{2}{c|}{Novel} \\ \cline{2-9} 
                        & AP & \multicolumn{1}{c|}{AP50} & AP          & AP50         & AP & \multicolumn{1}{c|}{AP50} & AP          & AP50   \\ \hline
Linear+Linear &
  \textbf{44.77} &
  \multicolumn{1}{c|}{\textbf{62.40}} &
  13.26 &
  18.54 &
  \textbf{41.60} &
  \multicolumn{1}{c|}{\textbf{60.88}} &
  12.25 &
  18.03 \\ \hline
Binary+Linear &
  {\ul{44.19}} &
  \multicolumn{1}{c|}{{\ul{61.58}}} &
  {\ul{14.41}} &
  {\ul{20.12}} &
  {\ul{39.97}} &
  \multicolumn{1}{c|}{{\ul{60.02}}} &
  {\ul{13.28}} &
  {\ul{19.62}} \\ \hline
Binary+Cosine &
  42.13 &
  \multicolumn{1}{c|}{59.43} &
  \textbf{15.59} &
  \textbf{22.15} &
  38.19 &
  \multicolumn{1}{c|}{57.91} &
  \textbf{14.33} &
  \textbf{21.62} \\ \hline
\end{tabular}
  }
 \vspace{-0.2cm}
    \label{tab:classifiers}
\end{table}

\begin{table*}[]
  \caption{\normalsize  Comparison of iFSOD and iFSIS with $k$=10 for different components on COCO 2014. We report AP and AP50 for base and novel classes. Optimal results are in \textbf{bold}, sub-optimal best results are in \ul{underline}. }
  \centering
  \renewcommand{\arraystretch}{1.3} %
  \resizebox{0.7\columnwidth}{!}{
  \begin{tabular}{|c|c|cccc|cccc|}
\hline
\multirow{3}{*}{\begin{tabular}[c]{@{}c@{}}Base\\ Fine-tuning\end{tabular}} &
  \multirow{3}{*}{\begin{tabular}[c]{@{}c@{}}Knowledge\\ Distillation\end{tabular}} &
  \multicolumn{4}{c|}{Object Detection} &
  \multicolumn{4}{c|}{Instance Segmentation} \\ \cline{3-10} 
 &
   &
  \multicolumn{2}{c|}{Base} &
  \multicolumn{2}{c|}{Novel} &
  \multicolumn{2}{c|}{Base} &
  \multicolumn{2}{c|}{Novel} \\ \cline{3-10} 
 &
   &
  AP &
  \multicolumn{1}{c|}{AP50} &
  AP &
  AP50 &
  AP &
  \multicolumn{1}{c|}{AP50} &
  AP &
  AP50 \\ \hline
\multicolumn{1}{|l|}{} &
  \multicolumn{1}{l|}{} &
  29.64 &
  \multicolumn{1}{c|}{43.39} &
  {\ul{16.39}} &
  {\ul{24.02}} &
  26.87 &
  \multicolumn{1}{c|}{42.00} &
  {\ul{14.65}} &
  {\ul{23.00}} \\ \hline
\checkmark &
   &
  27.49 &
  \multicolumn{1}{c|}{40.81} &
  \textbf{16.66} &
  \textbf{24.42} &
  24.83 &
  \multicolumn{1}{c|}{39.32} &
  \textbf{14.86} &
  \textbf{23.41} \\ \hline
 &
  \checkmark &
  \textbf{44.50} &
  \multicolumn{1}{c|}{\textbf{62.04}} &
  14.13 &
  20.07 &
  \textbf{40.28} &
  \multicolumn{1}{c|}{\textbf{60.71}} &
  13.07 &
  19.69 \\ \hline
\checkmark &
  \checkmark &
  {\ul{42.13}} &
  \multicolumn{1}{c|}{{\ul{59.43}}} &
  15.59 &
  22.15 &
  {\ul{38.19}} &
  \multicolumn{1}{c|}{{\ul{57.91}}} &
  14.33 &
  21.62 \\ \hline
\end{tabular}
  }
 \vspace{-0.2cm}
    \label{tab:components}
\end{table*}

\subsubsection{Dataset and Evaluation Metrics}We evaluate the proposed UIFormer on one popular benchmark datasets COCO 2014 \cite{lin2014microsoft} in the settings of both iFSIS and iFSOD. We adopted the AP (average precision) and AP50 (average precision at IoU=0.5) as our standard evaluation metrics. Following \cite{nguyen2022ifs}, we employed the 20 shared classes of COCO 2014 as the novel classes, and the remaining 60 classes as the base classes. 80k train and 35k validation images of this union dataset are used for training, and the remaining 5k validation images are used for testing. We evaluated the performance of the algorithms by varying the number of shots per novel class $k$, which is set to 1,5,10. We report average results of all test over 10 runs with $k$ random examples per class.

The LVIS dataset is a dataset with a long tail property that contains 1230 categories. According to the appearance frequency of each category in images,  \cite{nguyen2022ifs} divides these categories into three classes: rare classes (1-10 images), common classes (11-100 images) and frequent classes (\textgreater 100 images). We regard common classes and frequent classes in LVIS as base classes, and rare classes as novel classes.

\subsubsection{Competitors} We compare the performance of UIFormer with the three well known iFSIS and iFSOD methods: iMTFA\cite{ganea2021incremental}, iFS-RCNN\cite{nguyen2022ifs} and RefT\cite{reft2024}. iMTFA is the first paper to support both iFSIS and iFSOD simultaneously. iFS-RCNN, as the current SOTA, first proposes to use the Mask-R-CNN architecture to unify iFSIS and iFSOD. RefT addresses the FSIS problem based on Mask2Former and extends the model to the iFSIS setting.

\subsubsection{Implementation Details} Our UIFormer uses ResNet-50 as CNN backbone to extract the multi-scale features from original images. The network architectures and hyper-parameters of transformer encoder and decoder remain the same as Mask DINO\cite{li2023mask}. We used 2 RTX 3090 GPUs for training. The batch size of each GPU is set to 3 during base pre-training, and set to 4 during base fine-tuning and novel fine-tuning. We used the Adam optimizer for training, with an initial learning rate of 0.0001 during base pre-training. The number of iterations for base pre-training stage is 378750 with two weight decay steps with rate of 0.05 at 327778 and 355092 iterations. In base fine-tuning, the learning rate is set to 0.00005, and the number of iterations is set to 75000. In the novel fine-tuning, we trained the model until convergence. We set the maximum detections per image to 100 for COCO 2014 and  the number of object queries to 300 on both datasets.

\subsection{Ablation Studies}
\subsubsection{Effectiveness of different classification heads combination strategies.}
The following ablations are evaluated on COCO 2014 for studying how different classification heads combination strategies of UIFormer affects final performance.
\begin{itemize}
    \item \textbf{Linear+Linear} UIFormer uses linear classifiers as encoder and decoder’s classification heads.
    \item \textbf{Binary+Linear} UIFormer uses binary classifiers as encoder’s classification heads for foreground prediction and linear classifiers as decoder’s classification heads.
    \item \textbf{Binary+Cosine} UIFormer uses binary classifiers as encoder’s classification heads for foreground prediction and cosine classifiers as decoder’s classification heads.
\end{itemize}

Table~\ref{tab:classifiers} shows our evaluation of the above ablations
 on iFSOD and iFSIS. Compared with the model using the same classifier, the models using different classifiers effectively improves the performance of the novel classes with a slight decrease that of the base classes. More specifically, compared with Linear+Linear, for iFSOD, the performance of Binary+Linear and Binary+Cosine dropped by -0.58 (1.3\%) AP and -2.64 (5.9\%) AP for base classes, but they achieve an improvement of +1.15 (8.67\%) AP and +2.33 (17.57\%) AP for novel classes; for iFSIS, the performance of Binary+Linear and Binary+Cosine decreased by -1.63 (3.91\%) AP and -3.41 (8.20\%) AP for base classes, but they obtain an improvement of +1.03 (8.41\%) AP and +2.08(16.98\%) AP for novel classes.  Compared with Binary+Linear, Binary+Cosine performs better on novel classes, which is a more important for incremental few-shot learning. Thus, we choose Binary+Cosine as the default classification heads combination strategies.

\subsubsection{Effectiveness of different components.}Table~\ref{tab:components}  shows the ablation studies to evaluate the effectiveness of base fine-tuning with attention-driven pseudo ground-truth and knowledge distillation strategies in our framework on iFSIS and iFSOD. The experimental results demonstrate that compared with the model without base fine-tuning, the model using base fine-tuning can effectively improve the performance of novel class when the performance of base class decreases slightly. More Specifically, for iFSOD, if both models use knowledge distillation, the model using base fine-tuning meets a decrease of -2.37 (5.3\%) AP and -2.61 (4.21\%) AP50 for base classes, but it achieves an improvement of +1.46 (10.33\%) AP and +2.08 (10.36\%) AP50 for novel classes; for iFSIS, the model using base fine-tuning meets a decrease of -2.09 (5.18\%) AP and - 2.8 (4.61\%) AP50 for base classes, but it obtains an improvement of +1.26 (9.64\%) AP and +1.93 (9.80\%) AP50 for novel classes. 

Experimental results also show that the model using knowledge distillation achieves better performance than that without knowledge distillation for base classes, which indicates that knowledge distillation can effectively alleviate the catastrophic forgetting caused by the transformer-based model. More Specifically, for iFSOD, if both models take base fine-tuning strategy, the model using knowledge distillation achieves an improvement of +14.64 (53.26\%) AP and +18.62 (45.63\%) AP50 for novel classes; for iFSIS, the model using knowledge distillation has an increase of +13.36 (53.81\%) AP and +18.59 (47.28\%) AP50 for novel classes.

\subsection{Results on COCO}
We compare our UIFormer against iFS-RCNN, RefT and iMTFA on the base classes and novel classes. We report iFSIS and iFSOD performance on the COCO 2014 with different $k$ in Table~\ref{tab:coco2014}. 

\subsubsection{Results on iFSIS} 

For all methods, the performance of iFSIS increases with the number of shots for all class. UIFormer,iFS-RCNN and RefT outperform iMTFA by a large margin in terms of AP and AP50, for every number of tested shots. Our UIFormer significantly outperforms iFS-RCNN and RefT on all, base and novel classes, except for 1-shot setting of base and all classes, where all methods meet poor performance owing to the scarcity of data. More specifically, compared with iFS-RCNN, when $k$=5, our performance gains are about +1.79 AP for the novel classes and +2.76 AP for the base classes; when $k$=10, our performance gains are about +4.27 AP for the novel classes and +1.87 AP for the base classes. 

\begin{table}[]
  \caption{\normalsize  Comparison of iFSOD and iFSIS with AP metric on LVIS. Optimal results are in \textbf{bold}, sub-optimal best results are in \ul{underline}. Base-c and Base-f indicate the common classes (11-100 images) and frequent classes (\textgreater 100 images) among the base classes}
  \centering
  \renewcommand{\arraystretch}{1.3} %
  \resizebox{0.7\columnwidth}{!}{
\begin{tabular}{|c|ccc|ccc|}
\hline
\multirow{2}{*}{Methods} & \multicolumn{3}{c|}{Object Detection}      & \multicolumn{3}{c|}{Instance Segmentation} \\ \cline{2-7} 
         & Novel & Base-c         & Base-f & Novel & Base-c         & Base-f \\ \hline
iFS-RCNN & 18.38 & \textbf{26.11} & 30.12  & 18.26 & \textbf{26.29} & 28.46  \\ \hline
UIFormer(Ours)           & \textbf{22.456} & 16.975 & \textbf{32.338} & \textbf{21.738} & 17.584 & \textbf{31.816} \\ \hline
\end{tabular}
  }
 \vspace{-0.2cm}
    \label{tab:LVIS}
\end{table}

\subsubsection{Results on iFSOD}
Similar to iFSIS, the performance of iFSOD increases with the number of shots for all class.  UIFormer,iFS-RCNN and RefT outperform iMTFA by a large margin in terms of AP and AP50, when the number of tested shots is larger than 1. Our UIFormer outperforms iFS-RCNN and RefT on all, base and novel classes, except for 5-shot setting of novel classes, when tested shots is larger than 1. More specifically, compared with iFS-RCNN, when $k$=5, our performance gains are about +3.06 AP for the base classes and +2.66 AP for the novel classes; when $k$=10, our performance gains are about +2.08 AP for the base classes and +1.35 AP for the novel classes. 

\begin{table*}[]
  \caption{\normalsize  Comparison of iFSOD and iFSIS with different $k$={1,5,10} on COCO 2014. We report AP and AP50 for all, base and novel classes. Optimal results are in \textbf{bold}, sub-optimal best results are in \ul{underline}. }
  \centering
  \renewcommand{\arraystretch}{1.3} %
  \resizebox{\textwidth}{!}{
\begin{tabular}{|c|c|cccccc|cccccc|}
\hline
\multirow{3}{*}{Shots} & \multirow{3}{*}{Methods} & \multicolumn{6}{c|}{Object Detection}                                                                                                         & \multicolumn{6}{c|}{Instance Segmentation}                                                                                                    \\ \cline{3-14} 
                       &                          & \multicolumn{2}{c|}{All}                             & \multicolumn{2}{c|}{Base}                            & \multicolumn{2}{c|}{Novel}      & \multicolumn{2}{c|}{All}                             & \multicolumn{2}{c|}{Base}                            & \multicolumn{2}{c|}{Novel}      \\ \cline{3-14} 
                       &                          & AP             & \multicolumn{1}{c|}{AP50}           & AP             & \multicolumn{1}{c|}{AP50}           & AP             & AP50           & AP             & \multicolumn{1}{c|}{AP50}           & AP             & \multicolumn{1}{c|}{AP50}           & AP             & AP50           \\ \hline
\multirow{4}{*}{1}     & iMTFA                    & 21.67          & \multicolumn{1}{c|}{31.55}          & 27.81          & \multicolumn{1}{c|}{40.11}          & 3.23           & 5.89           & 20.13          & \multicolumn{1}{c|}{30.64}          & 25.9           & \multicolumn{1}{c|}{39.28}          & 2.81           & 4.72           \\ \cline{2-14} 
                       & iFS-RCNN                 & {\ul{31.19}}    & \multicolumn{1}{c|}{{\ul{45.52}}}    & {\ul{40.08}}    & \multicolumn{1}{c|}{{\ul{58.13}}}    & \textbf{4.54}  & \textbf{7.69}  & {\ul{28.45}}    & \multicolumn{1}{c|}{{\ul{43.73}}}    & 36.35          & \multicolumn{1}{c|}{{\ul{56.08}}}    & \textbf{3.95}  & \textbf{6.68}  \\ \cline{2-14} 
                       & RefT                     & -              & \multicolumn{1}{c|}{-}              & 32.20          & \multicolumn{1}{c|}{-}              & {\ul{4.00}}     & -              & -              & \multicolumn{1}{c|}{-}              & {\ul{37.00}}    & \multicolumn{1}{c|}{-}              & 3.10           & -              \\ \cline{2-14} 
                       & UIFormer(Ours)           & \textbf{34.63} & \multicolumn{1}{c|}{\textbf{48.74}} & \textbf{44.90} & \multicolumn{1}{c|}{\textbf{63.21}} & 3.83           & {\ul{5.34}}     & \textbf{31.38} & \multicolumn{1}{c|}{\textbf{47.72}} & \textbf{40.65} & \multicolumn{1}{c|}{\textbf{61.88}} & {\ul{3.56}}     & {\ul{5.25}}     \\ \hline
\multirow{4}{*}{5}     & iMTFA                    & 19.62          & \multicolumn{1}{c|}{28.06}          & 24.13          & \multicolumn{1}{c|}{33.69}          & 6.07           & 11.15          & 18.22          & \multicolumn{1}{c|}{27.1}           & 22.56          & \multicolumn{1}{c|}{33.25}          & 5.19           & 8.65           \\ \cline{2-14} 
                       & iFS-RCNN                 & {\ul{32.52}}    & \multicolumn{1}{c|}{{\ul{47.74}}}    & {\ul{40.06}}    & \multicolumn{1}{c|}{{\ul{58.09}}}    & 9.91           & \textbf{16.71} & {\ul{29.89}}    & \multicolumn{1}{c|}{{\ul{45.81}}}    & {\ul{36.33}}    & \multicolumn{1}{c|}{{\ul{56.03}}}    & {\ul{8.80}}     & {\ul{15.15}}    \\ \cline{2-14} 
                       & RefT                     & -              & \multicolumn{1}{c|}{-}              & 31.80          & \multicolumn{1}{c|}{-}              & \textbf{12.00} & -              & -              & \multicolumn{1}{c|}{-}              & 35.30          & \multicolumn{1}{c|}{-}              & {\ul{8.80}}     & -              \\ \cline{2-14} 
                       & UIFormer(Ours)           & \textbf{35.24} & \multicolumn{1}{c|}{\textbf{49.68}} & \textbf{43.12} & \multicolumn{1}{c|}{\textbf{60.75}} & {\ul{11.62}}    & {\ul{16.46}}    & \textbf{31.97} & \multicolumn{1}{c|}{\textbf{48.5}}  & \textbf{39.09} & \multicolumn{1}{c|}{\textbf{59.32}} & \textbf{10.59} & \textbf{16.04} \\ \hline
\multirow{4}{*}{10}    & iMTFA                    & 19.26          & \multicolumn{1}{c|}{27.49}          & 23.36          & \multicolumn{1}{c|}{32.41}          & 6.97           & 12.72          & 17.87          & \multicolumn{1}{c|}{26.46}          & 21.87          & \multicolumn{1}{c|}{32.01}          & 5.88           & 9.81           \\ \cline{2-14} 
                       & iFS-RCNN                 & {\ul{33.02}}    & \multicolumn{1}{c|}{{\ul{48.57}}}    & {\ul{40.05}}    & \multicolumn{1}{c|}{{\ul{58.08}}}    & 12.55          & {\ul{20.04}}    & {\ul{30.41}}    & \multicolumn{1}{c|}{{\ul{46.59}}}    & {\ul{36.32}}    & \multicolumn{1}{c|}{{\ul{56.03}}}    & 10.06          & {\ul{18.29}}    \\ \cline{2-14} 
                       & RefT                     & -              & \multicolumn{1}{c|}{-}              & 33.40          & \multicolumn{1}{c|}{-}              & {\ul{14.90}}    & -              & -              & \multicolumn{1}{c|}{-}              & 35.20          & \multicolumn{1}{c|}{-}              & {\ul{11.10}}    & -              \\ \cline{2-14} 
                       & UIFormer(Ours)           & \textbf{35.49} & \multicolumn{1}{c|}{\textbf{50.11}} & \textbf{42.13} & \multicolumn{1}{c|}{\textbf{59.43}} & \textbf{15.59} & \textbf{22.15} & \textbf{32.23} & \multicolumn{1}{c|}{\textbf{48.83}} & \textbf{38.19} & \multicolumn{1}{c|}{\textbf{57.91}} & \textbf{14.33} & \textbf{21.62} \\ \hline
\end{tabular}
}

 \vspace{-0.2cm}
    \label{tab:coco2014}
\end{table*}

\subsubsection{Results Analysis}
In the 1-shot setting, the performance of iFS-RCNN (SOTA) is better than that of our UIFormer, because compared with the CNN-based model, the Transformer-based model requires more data for fine-tuning. But when the sample size is sufficient, the performance of our algorithm will be better than the CNN-based model. The experimental results of 5-shot setting and 10-shot setting on iFSIS and iFSOD also prove this view.

The difference between the performance of iFS-RCNN on iFSOD and iFSIS is significant and continues to expand as the number of tested slots increases. For example, for 5-shot setting, the difference between iFSIS and iFSOD on AP and AP50 in novel class is 1.11 and 1.56 respectively. However, these differences expand to 2.49 and 1.75 respectively. The difference between the performance of iFS-RCNN on iFSOD and iFSIS is trivial and just slightly expand as the number of tested slots increases. For example, for 5-shot setting, the difference between iFSIS and iFSOD on AP and AP50 in novel class is 1.03 and 0.42 respectively. However, these differences expand to 1.26 and 0.53 respectively. The experimental results indicate that the Transformer based model is a more effective unified framework compared to the CNN based model.

 \begin{figure*}[tb]
    \centering
    \includegraphics[height=5.2cm]{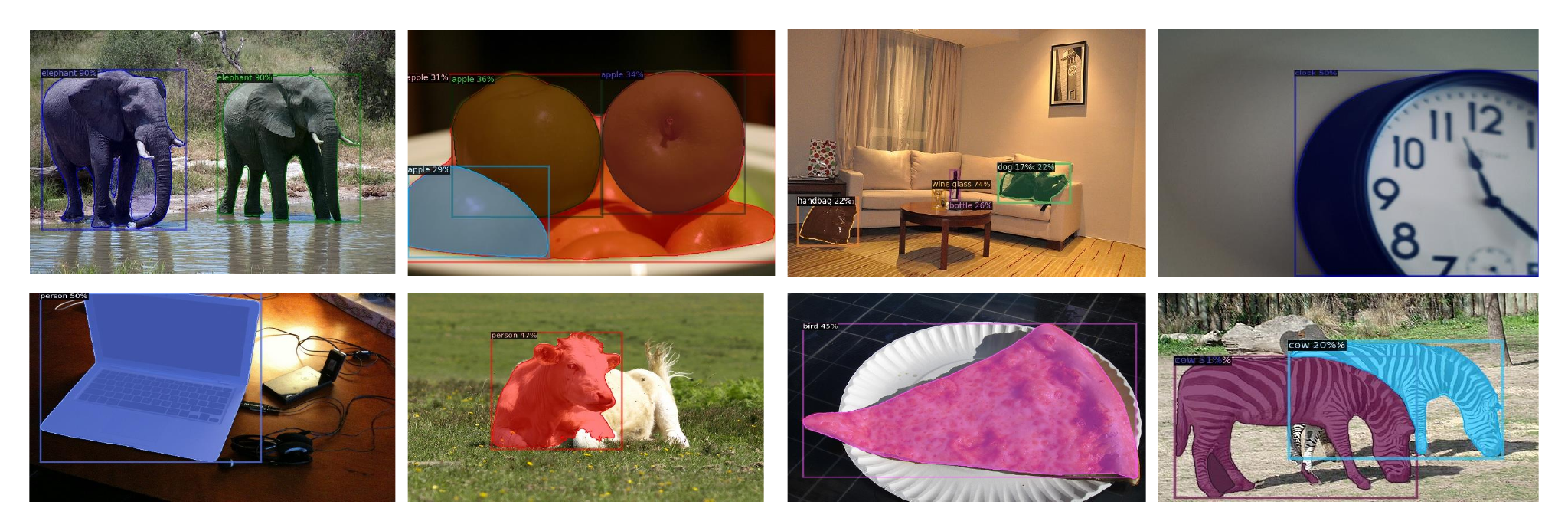}
    \caption{The visualization of representative results of our method on COCO 2014 with $k$=10. The top rows shows the success cases while the bottom row shows the failure cases.
    }
    \label{fig:vis}
    \vspace{-0.2cm}
\end{figure*}

\subsection{Results on LVIS}
Table~\ref{tab:LVIS}  reports our results on LVIS with $k$=10 on iFSOD and iFSIS. Compared with iFS-RCNN (SOTA), UIFormer significantly outperforms iFS-RCNN with the gains of +4.076 on iFSOD and +3.478 on iFSIS on the new classes, and +2.218 on iFSOD and +3.356 on iFSIS on the frequent classes among the base classes. The performance of UIFormer in common classes is inferior to that of CNN-based models because the number of images in common classes is far less than in Frequence class, so that it cannot provide sufficient training data for transformer-based models.

\subsection{Qualitative Evaluation}
Fig.\ref{fig:vis} presents a visualization of our results. The top two rows represent successful cases, while the bottom row shows the failure cases. Specifically, in the failed cases, the laptop is misclassified as person, the cow is misclassified as person, the pizza is misclassified as bird, and the zebra is misclassified as cow.

\section{Conclusion}
We propose a unified transformer-based framework for iFSOD and iFSIS, termed UIFormer. This novel two-stage learning framework extends Mask DINO by addressing its limitations. To mitigate over-fitting during base class learning, we introduce an attention-driven pseudo ground-truth search method that effectively captures rich semantics from the base classes. Additionally, we replace the original classifier with a novel class-agnostic foreground predictor, which further enhancing the foregrounds recognition on novel data. To further preserve base knowledge and prevent catastrophic forgetting, we incorporate knowledge distillation across base model and novel model. Extensive experiments conducted on the COCO dataset demonstrate the impressive performance of UIFormer in both iFSOD and iFSIS task.

\bibliographystyle{unsrt}  
\bibliography{references}

\end{document}